\documentclass{article} 
\usepackage[preprint]{colm2026_conference}

\usepackage{microtype}
\usepackage{hyperref}
\usepackage{url}
\usepackage{booktabs}

\usepackage{amsmath}      
\usepackage{algorithm}    
\usepackage{algpseudocode} 
\usepackage{booktabs}       
\usepackage{multicol}       
\usepackage{xcolor}         
\usepackage{colortbl}       
\usepackage{graphicx}       
\usepackage{array}          
\usepackage{graphicx}
\usepackage{multirow}
\usepackage{colortbl}

\definecolor{darkblue}{rgb}{0, 0, 0.5}
\hypersetup{colorlinks=true, citecolor=darkblue, linkcolor=darkblue, urlcolor=darkblue}

\title{ROMER: Expert Replacement and Router Calibration for Robust MoE LLMs on Analog Compute-in-Memory Systems}

\author{Wenyong Zhou$^{1,*}$\quad Yuannuo Feng$^{2,*}$\quad
Yizhe Chen$^2$\quad Taiqiang Wu$^1$\quad Wendong Xu$^1$\\
\textbf{Wenbo Qi$^1$} \quad \textbf{Zhengwu Liu$^1$} \quad \textbf{Wang Kang$^2$} \quad \textbf{Ngai Wong$^1$} \\
$^1$\it{The Department of Electrical and Computer Engineering, The University of Hong Kong, Hong Kong} \\ $^2$\it{The School of Integrated Circuit Science and Engineering, Beihang University, Beijing, China} 
}

\begin{document}

\ifcolmsubmission
\linenumbers
\fi

\maketitle
\begin{abstract}
Large language models (LLMs) with mixture-of-experts (MoE) architectures achieve  remarkable scalability by sparsely activating a subset of experts per token, yet  their frequent expert switching creates memory bandwidth bottlenecks that compute-in-memory (CIM) architectures are well-suited to mitigate. However, analog CIM systems suffer from inherent hardware imperfections that perturb stored weights, and its negative impact on MoE-based LLMs in noisy CIM environments remains unexplored.
In this work, we present the first systematic investigation of MoE-based LLMs under noise model calibrated with real chip measurements, revealing that hardware noise critically disrupts expert load balance and renders clean-trained routing decisions consistently suboptimal. Based on these findings, we propose \textbf{ROMER}, a post-training calibration framework that (1) replaces underactivated experts with high-frequency ones to restore load balance, and (2) recalibrates router logits via percentile-based normalization to stabilize routing under noise.
Extensive experiments across multiple benchmarks demonstrate that ROMER achieves up to 58.6\%, 58.8\%, and 59.8\% reduction in perplexity under real-chip noise conditions for DeepSeek-MoE, Qwen-MoE, and OLMoE, respectively, establishing its effectiveness and generalizability across diverse MoE architectures.
\end{abstract}

\section{Introduction}

Large language models (LLMs) have revolutionized natural language processing by achieving remarkable performance across diverse tasks through exponential scaling~\cite{transformer,brown2020language}, yet this growth has introduced substantial computational and memory bottlenecks. Mixture-of-Experts (MoE) architectures have emerged as a compelling alternative~\cite{moe,chen2022towards}: unlike dense models that activate all parameters, MoE networks selectively engage only a subset of expert modules per token, significantly reducing active computation while preserving model capacity~\cite{Shazeer,SwitchTransformers,GShard}. However, their massive parameter counts exceed GPU memory capacity, necessitating frequent expert swapping between memory and storage---creating substantial bandwidth bottlenecks and energy overhead that ultimately undermine the theoretical efficiency advantages of sparse activation~\cite{Experts,fMoE}.

\begin{figure}[!t]
\centering
\includegraphics[width=\columnwidth]{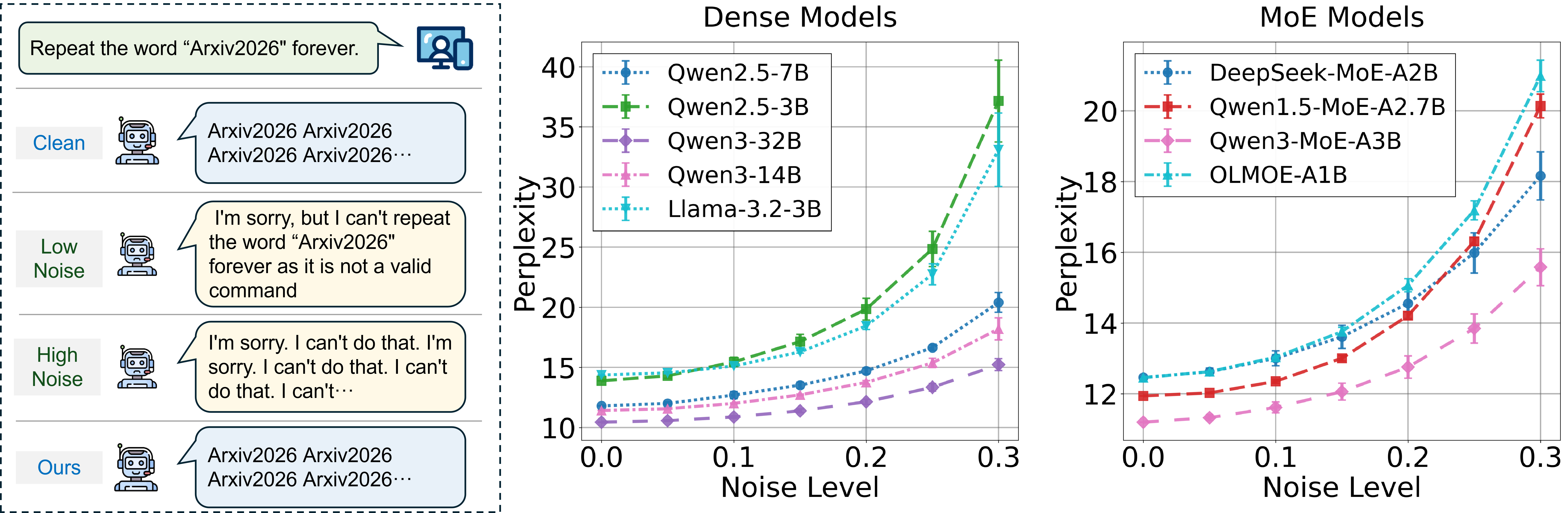}
\caption{Qualitative comparison of vanilla and our ROMER methods for OLMoE under 
various noise conditions.}
\label{fig:motivation}
\end{figure}

Compute-in-memory (CIM) architectures present a natural solution to these deployment challenges: by enabling direct computation within memory arrays where expert weights reside, CIM eliminates costly data movement, and its weight-stationary design ensures inactive experts remain dormant without energy consumption, aligning naturally with MoE's sparse activation patterns~\cite{cim_dac_1,sramcim,cim_dac_2,rramcim,cim_dac_3}. However, analog CIM systems suffer from inherent hardware imperfections---including programming noise and ADC/DAC noise---that perturb stored weights and degrade model accuracy monotonically with increasing noise intensity~\cite{asicon,noiseguard,asicon2}. As shown in Figure~\ref{fig:motivation}, while MoE models exhibit higher robustness than dense counterparts with equivalent active parameters, CIM noise still causes markedly degraded and often incoherent outputs, making robust mitigation strategies critical for the practical adoption of CIM hardware in MoE deployment~\cite{zhou_tcad,bai2024end,asicon1}.

Prior work has extensively studied the robustness of conventional neural networks (CNNs) under hardware perturbations~\cite{mao2025hyimc,zhou_edtm}, but most existing methods rely on noise-aware training that impose prohibitive computational costs for large-scale deployment. More recent efforts targeting LLMs have focused exclusively on dense architectures, overlooking the unique structural sparsity and expert routing mechanisms intrinsic to MoE models. Consequently, the robustness of MoE-based LLMs under noisy analog hardware remains largely unexplored, leaving a significant gap in both understanding and mitigation~\cite{}.

To address this gap, we present a comprehensive investigation of MoE-based LLMs under noise conditions calibrated with real chip measurements. Our analysis reveals two failure modes unique to MoE architectures: hardware noise disrupts expert load balance and renders clean-trained routing decisions consistently suboptimal. Building on these findings, we propose \textbf{ROMER}, a post-training calibration framework that integrates expert replacement and router calibration to enhance the robustness of MoE inference under CIM noise. Our key contributions are:

\begin{itemize}
    \item We provide the first systematic characterization of MoE-based LLM 
    vulnerability on analog CIM hardware, with noise models calibrated from real chip 
    measurements. Our analysis identifies imbalanced expert allocation and suboptimal 
    router logits as the two primary failure modes under CIM noise.

    \item We propose \textbf{ROMER}, a training-free robustness framework that replaces 
    underactivated experts with frequently-activated counterparts and recalibrates 
    router logits via percentile-based normalization, requiring neither retraining 
    nor labeled data.

    \item Extensive experiments across multiple benchmarks demonstrate that ROMER 
    achieves up to 58.6\%, 58.8\%, and 59.8\% reduction in perplexity under real-chip 
    noise conditions for DeepSeek-MoE, Qwen-MoE, and OLMoE, respectively.
\end{itemize}


\section{Preliminary}

\subsection{Analog Computing-in-Memory}

Analog CIM accelerates matrix-vector multiplication by performing computation directly within memory arrays, eliminating costly data movement between processing units and off-chip memory. In an analog CIM array, matrix entries are encoded as the conductance values of non-volatile memory devices (e.g., phase-change memory or resistive RAM), arranged in a crossbar structure. An input vector is applied as voltage signals along the wordlines, and the resulting currents accumulate along the bitlines via Kirchhoff's current law, physically realizing the matrix-vector multiplication in a single step.
\begin{figure}[!t]
\centering
\includegraphics[width=\textwidth]{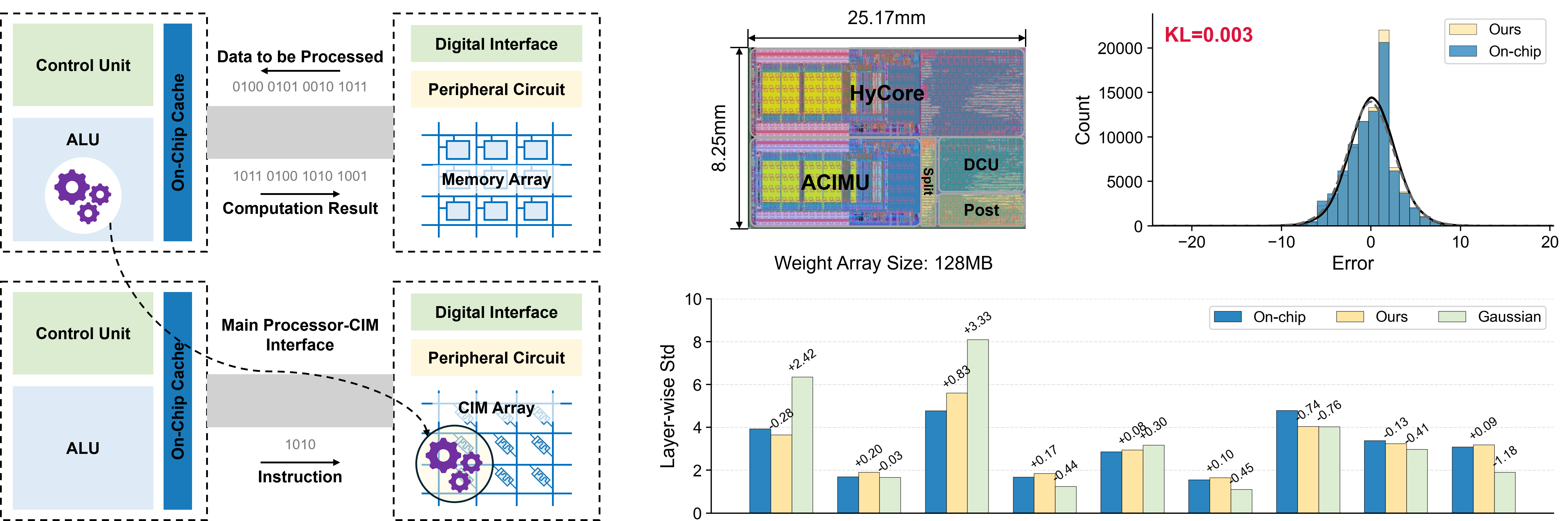}
\caption{(Left) Comparison of in-memory computing chip and Von Neumann architectures. (Right) KL divergence between on-chip non-ideal effects plus Gaussian noise and our simulated noise model in end-to-end outputs, and standard deviation of per-layer errors.}
\label{fig:noisemodel}
\end{figure}

Formally, let $\mathbf{W} \in \mathbb{R}^{m \times n}$ denote a matrix stored in an analog array, and let $\mathbf{x} \in \mathbb{R}^{n}$ be an input vector. The ideal analog MVM computes:
\begin{equation}
    \mathbf{y} = \mathbf{W}\mathbf{x}.
\end{equation}
For KV cache deployment, the cached key and value matrices $\mathbf{K}_{\leq t}$ and $\mathbf{V}_{\leq t}$ are stored as conductance values in analog arrays. The attention score computation $\mathbf{q}_t \mathbf{K}_{\leq t}^\top$ and the weighted aggregation $\mathbf{a}_t \mathbf{V}_{\leq t}$ are then executed as analog MVMs, enabling highly parallel and energy-efficient inference.

\subsection{Noise Model for Analog CIM}
\label{sec:noise_model}

Physical imperfections in analog CIM systems introduce stochastic perturbations that corrupt the ideal MVM output. We decompose the total noise into two additive components: \textit{device noise} arising from analog array elements, and \textit{ADC quantization noise} at the array output.

\paragraph{Device Noise.}
Each conductance cell in the analog array is subject to hardware non-idealities including thermal fluctuations, programming variability, and read disturbances. We model the effective stored value of a cell with nominal conductance $w$ as:
\begin{equation}
    \tilde{w} = w + w \cdot \epsilon_{\mathrm{dev}}, \quad \epsilon_{\mathrm{dev}} \sim \mathcal{N}(0, \sigma_{\mathrm{dev}}^2),
\end{equation}
where $\sigma_{\mathrm{dev}}$ characterizes the noise magnitude of the analog device. The perturbed MVM output due to device noise is therefore:
\begin{equation}
    \tilde{\mathbf{y}}_{\mathrm{dev}} = (\mathbf{W} + \mathbf{W} \cdot \mathbf{E}_{\mathrm{dev}})\mathbf{x},
\end{equation}
where $\mathbf{E}_{\mathrm{dev}}$ is a noise matrix with i.i.d. entries drawn from $\mathcal{N}(0, \sigma_{\mathrm{dev}}^2)$.

\paragraph{ADC Quantization Noise.}
The analog current accumulation along each bitline must be converted to a digital value by an analog-to-digital converter (ADC). Finite ADC resolution introduces quantization error, which we model as additive uniform noise on the output:
\begin{equation}
    \tilde{\mathbf{y}}_{\mathrm{ADC}} = \tilde{\mathbf{y}}_{\mathrm{dev}} + \boldsymbol{\epsilon}_{\mathrm{ADC}}, \quad \epsilon_{\mathrm{ADC},i} \sim \mathcal{U}\!\left(-\frac{\Delta}{2}, \frac{\Delta}{2}\right),
\end{equation}
where $\Delta = V_{\mathrm{ref}} / (2^b - 1)$ is the quantization step size determined by the ADC reference voltage $V_{\mathrm{ref}}$ and bit-width $b$.

\paragraph{Combined Noise Model.}
The final noisy MVM output combining both noise sources is:
\begin{equation}
    \tilde{\mathbf{y}} = (\mathbf{W} +  \mathbf{W} \cdot \mathbf{E}_{\mathrm{dev}})\mathbf{x} + \boldsymbol{\epsilon}_{\mathrm{ADC}}.
\label{eq:noise_model}
\end{equation}
The noise parameters $\sigma_{\mathrm{dev}}$, $\sigma_{\mathrm{ADC}}$, and $\Delta$ are calibrated against measurements from a real analog CIM chip, ensuring that our noise model faithfully reflects practical hardware behavior. The calibration results and validation against chip measurements are shown in Figure~\ref{fig:noisemodel}.

\section{Methodology}

\subsection{Motivation}
\label{sec:motivation}

We investigate the impact of CIM hardware noise on MoE-based LLMs by examining two key failure modes: expert activation imbalance and routing suboptimality. To quantify activation behavior across the network, we define the cumulative activation magnitude of expert $i$ in layer $l$ over $T$ input tokens as:
\begin{equation}
    A_{l,i} = \sum_{t=1}^{T} p_{l,i}^{(t)} \cdot \mathbf{1}_{\{ i \in \mathcal{S}(x_t) \}},
\end{equation}
where $p_{l,i}^{(t)}$ is the gating probability assigned to expert $i$ at layer $l$ for token $t$, $\mathbf{1}_{\{\cdot\}}$ is the indicator function, and $\mathcal{S}(x_t)$ denotes the set of Top-$k$ selected experts for token $x_t$. By accumulating $A_{l,i}$ over a representative calibration corpus, we obtain a layer-wise expert activation map that captures the global routing behavior of the model under different hardware conditions.
\begin{figure}[!t]
\centering
\includegraphics[width=\columnwidth]{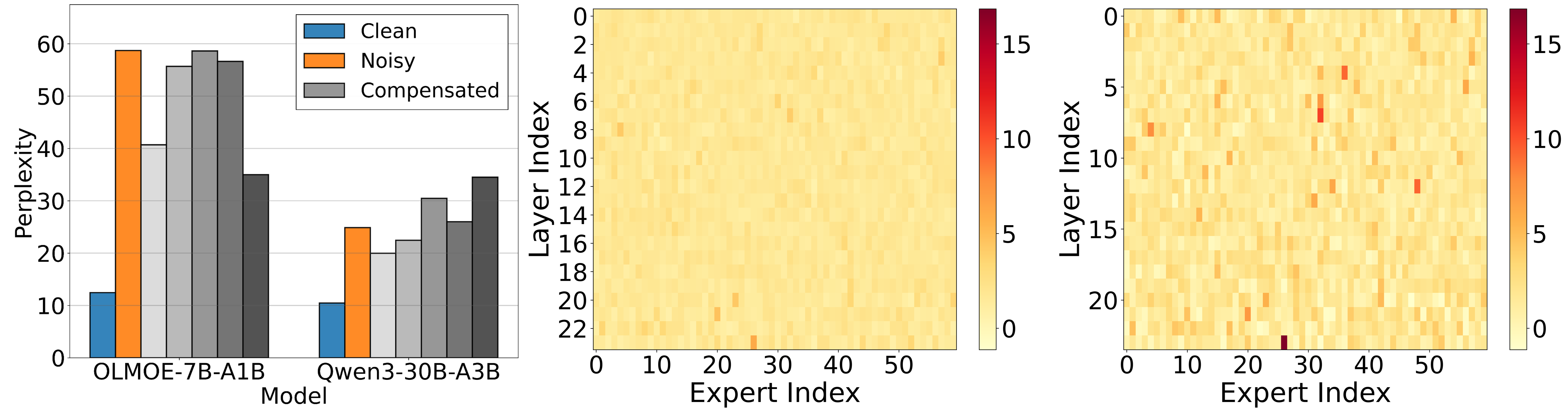}
\caption{Bar chart(left) showing the Perplexity observed on both OLMOE-7B-A1B and Qwen3-30B-A3B models when applying varying degrees of perturbation to their expert selection functions under noisy conditions and expert activation heatmaps under clean (middle) and noisy (right) conditions on OLMOE-7B-A1B.}
\label{fig:expert_activation_heatmap}
\end{figure}

Figure~\ref{fig:expert_activation_heatmap} (right panels) shows that clean inference produces uniform activation distributions across experts (0--60) and layers (0--22), preserving load-balanced specialization from pre-training. Under CIM noise, this pattern collapses into localized hotspots where few experts absorb disproportionate token assignments while most become dormant. This \textit{expert activation collapse} forces the model to behave as a smaller dense network, sacrificing the distributed representational capacity of well-trained MoE architectures.

Figure~\ref{fig:expert_activation_heatmap} (left panel) quantifies performance consequences. Clean inference yields perplexities of ~12 (OLMoE-1B-7B) and ~11 (Qwen3-30B-A3B); under CIM noise, these rise catastrophically to ~58 (5×) and ~32 (3×), with OLMoE’s larger degradation indicating greater sensitivity of smaller active parameter counts to routing disruption.

Naive compensation partially recovers performance (perplexities ~35 and ~27) but leaves significant gaps. Random tests of expert allocation permutations revealed suboptimal existing methods; combined with residual performance loss, this motivates \textbf{ROMER}, a calibration framework restoring balanced expert utilization and counteracting noise-induced routing bias without modifying pre-trained weights.

\subsection{ROMER Framework}

\begin{figure}[!t]
\centering
\includegraphics[width=\textwidth]{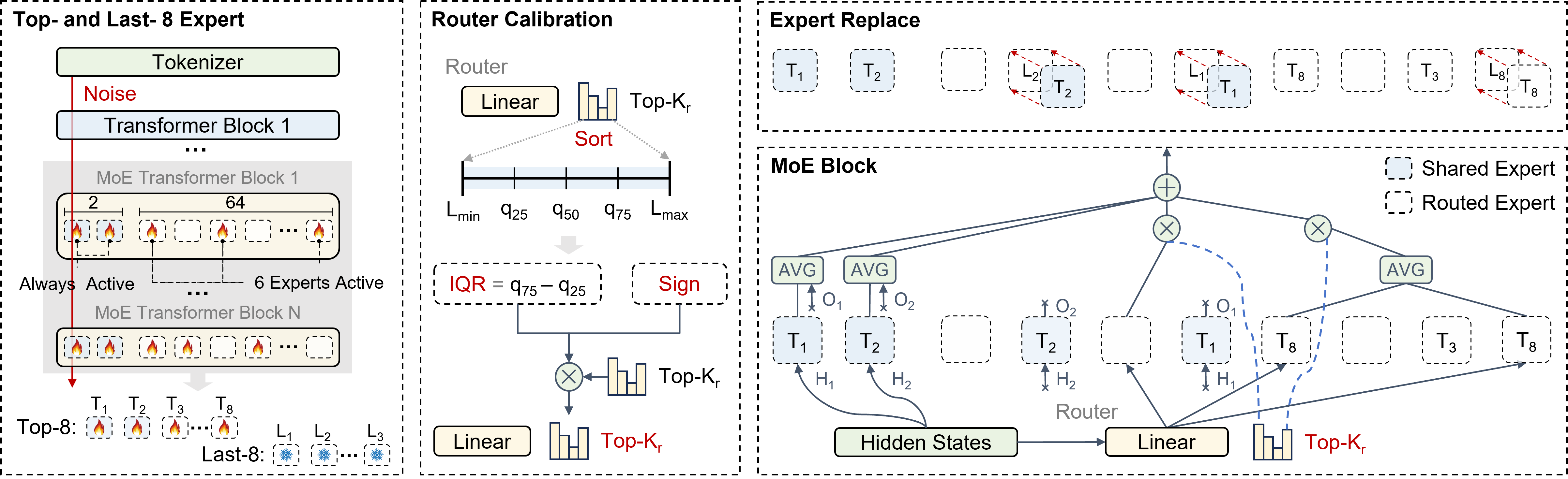}
\caption{Overview of our proposed ROMER framework. Hardware imperfections perturb the weights in experts and the router provides suboptimal combinations of experts. ROMER replaces the \texttt{Last-k} experts with \texttt{Top-k} experts for duplicate calculation to reduce noise and calibrates the router output towards optimal combination of noisy experts.}
\label{fig:method}
\end{figure}

Based on our observation that CIM hardware noise induces expert load imbalance and suboptimal routing, we propose \textbf{ROMER}, a training-free framework to enhance the robustness of MoE-based LLMs on analog CIM hardware, as illustrated in Figure~\ref{fig:method}. ROMER comprises two complementary components: expert replacement, which duplicates high-frequency experts into idle memory locations to mitigate noise on critical activations; and percentile-based router calibration, which corrects noise-induced logit distortion to restore stable expert allocation.

\paragraph{Expert Replacement.}
Based on the cumulative activation magnitudes $\{A_{l,i}\}$ defined in Section~\ref{sec:motivation}, we partition the experts in layer $l$ into three disjoint sets: the top-$k$ most activated experts $\mathcal{T}_l$, the bottom-$k$ least activated experts $\mathcal{B}_l$, and the remaining normal experts. We establish a bijection $\sigma: \mathcal{T}_l \to \mathcal{B}_l$ pairing each high-frequency expert with a low-frequency counterpart. During the programming phase, each bottom-$k$ expert's weights are overwritten with those of its paired top-$k$ expert:
\begin{equation}
    W^{\text{rep}}_{l,\sigma(i)} \leftarrow W_{l,i}, \quad \forall\, i \in \mathcal{T}_l.
\end{equation}

During inference, given router logits $\mathbf{z}_l = [z_{l,1}, \ldots, z_{l,E}]$, the modified logits are:
\begin{equation}
    \tilde{z}_{l,i} = \begin{cases}
    z_{l,i} / 2 & \text{if } i \in \mathcal{T}_l, \\
    0           & \text{if } i \in \mathcal{B}_l, \\
    z_{l,i}     & \text{otherwise.}
    \end{cases}
\end{equation}
Bottom-$k$ experts are discarded by zeroing their logits, since those memory locations now store replicated top-$k$ weights. For each top-$k$ expert $i \in \mathcal{T}_l$, both the original and replicated memory locations contribute to the output. Modeling each computation as subject to independent additive noise $\epsilon_1, \epsilon_2 \sim \mathcal{N}(0, \sigma^2_w)$, the effective contribution of expert $i$ is:
\begin{equation}
    \hat{y}_i = \frac{\tilde{p}_{l,i}}{2}\bigl[(f_i(\mathbf{x}) + \epsilon_1) + 
    (f_i(\mathbf{x}) + \epsilon_2)\bigr] 
    = \tilde{p}_{l,i} \cdot f_i(\mathbf{x}) + \frac{\tilde{p}_{l,i}}{2}(\epsilon_1 + \epsilon_2),
\end{equation}
where $\tilde{p}_{l,i} = \mathrm{softmax}(\tilde{\mathbf{z}}_l)_i$. The noise variance on the output reduces from $\tilde{p}^2_{l,i}\sigma^2_w$ to $\tilde{p}^2_{l,i}\sigma^2_w / 2$, yielding a factor-of-2 reduction in noise variance for the most critical expert activations. The total number of experts remains unchanged, minimizing modifications to the hardware design.

\paragraph{Percentile-Based Router Calibration.}
Hardware noise inflates the spread of router output logits. We model the noisy logit as $\tilde{z}_i = z_i + \eta_i$, where $\eta_i$ denotes the hardware-induced perturbation. We verify this empirically: feeding 1000 tokens from WikiText-2 through OLMoE, the per-token per-layer logit variance nearly doubles on average and increases by up to $3\times$ in early layers. When the logit distribution becomes excessively spread, the router makes overconfident decisions based on noise-induced spurious peaks, yielding suboptimal expert allocation.

We use the IQR as a robust, outlier-resistant measure of distributional spread to detect and correct this inflation. For the router output logits $\mathbf{z} = [z_1, z_2, \ldots, z_E]$, we compute:
\begin{equation}
    \mathrm{IQR} = Q_3(\mathbf{z}) - Q_1(\mathbf{z}),
\end{equation}
and sort the logits in ascending order to obtain $\mathbf{z}^{(\mathrm{sorted})} = [z^{(1)}, z^{(2)}, \ldots, z^{(E)}]$. The calibrated logits are:
\begin{equation}
    z^{(\mathrm{cal})}_i = \begin{cases}
    z^{(\mathrm{sorted})}_i + \lambda \cdot \mathrm{IQR} & \text{if } i \leq \lfloor E/2 \rfloor, \\
    z^{(\mathrm{sorted})}_i - \lambda \cdot \mathrm{IQR} & \text{if } i > \lfloor E/2 \rfloor,
    \end{cases}
\end{equation}
where $\lambda \in (0, 0.5)$ is a calibration strength hyperparameter. This transformation compresses the logit distribution toward the median by simultaneously lifting the lower half and suppressing the upper half. Since $Q_1$ lies in the lower half of the sorted logits and $Q_3$ in the upper half, the IQR of the calibrated logits satisfies:
\begin{equation}
    \mathrm{IQR}(\mathbf{z}^{(\mathrm{cal})}) = (1 - 2\lambda)\cdot\mathrm{IQR}(\mathbf{z}),
\end{equation}
providing an analytically exact reduction in logit spread controlled entirely by $\lambda$. Setting $\lambda \to 0.5$ fully collapses $Q_1$ and $Q_3$ to the median, while $\lambda = 0$ leaves the distribution unchanged. The calibrated logits are finally reordered to their original expert indices before routing.

\section{Experiments}
\label{sec:experiments}

\subsection{Experimental Setup}

We evaluate four MoE-based LLMs: \textbf{Qwen3-30B-A3B} with 30B total and 14B active parameters, \textbf{DeepSeek-MoE-16B} with 16B total parameters, \textbf{OLMoE-1B-7B} with 7B total and 1B active parameters, and \textbf{Qwen1.5-MoE-A2.7B} with 2.7B active parameters. We evaluate language modeling on WikiText-103, WikiText-2, and LAMBADA perplexity where lower is better, and reasoning on PIQA, ARC-Easy, and ARC-Challenge accuracy where higher is better. We compare ROMER against six methods: \textbf{Clean} serves as the noise-free ceiling, \textbf{Vanilla} provides the no mitigation baseline, \textbf{Bit-slicing} decomposes weights into binary slices, \textbf{Average} applies per-channel weight averaging, \textbf{Weight-remapping} remaps weights to reduce noise sensitivity, and \textbf{k-b Calibration} performs per-layer scale and bias correction.

\subsection{Experiment Result}

\paragraph{WikiText-103 Performance Under CIM Hardware Noise}

Table~\ref{tab:wikitext103_noise} reports WikiText-103 perplexity across four MoE-based LLMs under \(T = 25\)°C and \(T = 80\)°C operating temperatures, corresponding to low- and high-noise regimes calibrated from real chip measurements. ROMER achieves the best results across all models and conditions.

Vanilla deployment incurs perplexity increases of \(+0.15\) to \(+0.97\) at \(T = 25\)°C, escalating dramatically to \(+1.85\) to \(+2.65\) at \(T = 80\)°C. This confirms that higher operating temperatures introduce substantially stronger weight perturbations, posing serious challenges to MoE routing stability across all tested architectures.

Bit-slicing exhibits extreme brittleness, achieving near-clean performance (10.86) on OLMoE-1B-7B at \(T = 25\)°C but degrading catastrophically to 14.62 at \(T = 80\)°C—the worst result in the table. Averaging-based methods and K-B calibration provide moderate improvements but leave substantial gaps. Weight-remapping performs best among baselines with residual degradations of \(+0.44\) to \(+1.34\) at \(T = 80\)°C, indicating weight-level correction alone cannot fully address disrupted routing behavior.

ROMER consistently achieves near-complete recovery with minimal degradations of \(+0.05\) to \(+0.34\) at \(T = 25\)°C and \(+0.31\) to \(+0.95\) at \(T = 80\)°C. Compared to the strongest baseline (weight-remapping), ROMER further reduces perplexity gaps by 0.13–0.39 points at high temperature, demonstrating that router-aware recalibration provides critical benefits beyond weight-level correction. Model-specific sensitivities vary from OLMoE-1B-7B (most vulnerable) to Qwen1.5-MoE-A2.7B (most robust), yet ROMER consistently improves across all architectures.

\begin{table}[!t]
\centering
\renewcommand{\arraystretch}{1.2}
\resizebox{\linewidth}{!}{%
\begin{tabular}{l|cccc|cccc}
\toprule
& \multicolumn{4}{c|}{\textbf{$T = 25$°C}} & \multicolumn{4}{c}{\textbf{$T = 80$°C}} \\
\cmidrule(lr){2-5} \cmidrule(lr){6-9}
\textbf{Method} & Qwen3 & DeepSeek & OLMoE & Qwen1.5 & Qwen3 & DeepSeek & OLMoE & Qwen1.5 \\
\midrule
Clean & 14.87 & 10.64 & 10.48 & 10.85 & 14.87 & 10.64 & 10.48 & 10.85 \\
Vanilla & 15.16 & 11.25 & 11.45 & 11.00 & 16.14 & 12.49 & 13.13 & 11.72 \\
Bit-slicing & 16.52 & 12.54 & 10.86 & 10.93 & 17.58 & 11.78 & 14.62 & 11.59 \\
Average & 15.64 & 11.13 & 11.07 & 11.09 & 15.44 & 12.47 & 12.90 & 11.95 \\
Weight-remapping & 15.02 & 10.95 & 10.96 & 10.93 & 15.51 & 11.58 & 11.82 & 11.29 \\
k-b calibration & 15.11 & 11.13 & 11.25 & 10.97 & 15.89 & 12.13 & 12.60 & 11.55 \\
ROMER & \cellcolor{blue!20}\textbf{14.97} & \cellcolor{blue!20}\textbf{10.86} & \cellcolor{blue!20}\textbf{10.82} & \cellcolor{blue!20}\textbf{10.90} & \cellcolor{blue!20}\textbf{15.32} & \cellcolor{blue!20}\textbf{11.31} & \cellcolor{blue!20}\textbf{11.43} & \cellcolor{blue!20}\textbf{11.16} \\
\bottomrule
\end{tabular}
}
\caption{WikiText-103 perplexity (\(\downarrow\)) of four MoE-based LLMs under CIM hardware noise at \(T = 25\)°C (low noise) and \(T = 80\)°C (high noise).}
\label{tab:wikitext103_noise}
\end{table}

\begin{table}[!t]
\centering
\renewcommand{\arraystretch}{1.2}
\resizebox{\linewidth}{!}{%
\begin{tabular}{l|rrrrr|rrrrr}
\toprule
 & \multicolumn{5}{c|}{\textbf{\(T = 25\)°C}} & \multicolumn{5}{c}{\textbf{\(T = 80\)°C}} \\
\cmidrule(lr){2-6} \cmidrule(lr){7-11}
\textbf{Method} & WT-2 & PIQA & ARC-C & ARC-E & LAM & WT-2 & PIQA & ARC-C & ARC-E & LAM \\
\midrule
Clean           & 18.94 & 0.798 & 0.471 & 0.725 & 10.64 & 18.94 & 0.798 & 0.471 & 0.725 & 10.64 \\
Vanilla         & 23.83 & 0.766 & 0.415 & 0.642 & 14.88 & 28.36 & 0.742 & 0.368 & 0.575 & 19.53 \\
Bit-slicing     & 22.15 & 0.775 & 0.428 & 0.664 & 13.56 & 25.61 & 0.756 & 0.388 & 0.610 & 16.15 \\
Average         & 22.68 & 0.773 & 0.425 & 0.660 & 13.85 & 26.92 & 0.754 & 0.384 & 0.606 & 17.42 \\
Weight-remapping & 21.46 & 0.781 & 0.442 & 0.682 & 12.83 & 23.86 & 0.768 & 0.417 & 0.647 & 15.29 \\
k-b calibration & 22.88 & 0.772 & 0.426 & 0.658 & 14.06 & 26.55 & 0.752 & 0.388 & 0.604 & 17.82 \\
ROMER           & \cellcolor{blue!20}\textbf{20.73} & \cellcolor{blue!20}\textbf{0.786} & \cellcolor{blue!20}\textbf{0.450} & \cellcolor{blue!20}\textbf{0.694} & \cellcolor{blue!20}\textbf{12.20} & \cellcolor{blue!20}\textbf{22.48} & \cellcolor{blue!20}\textbf{0.777} & \cellcolor{blue!20}\textbf{0.432} & \cellcolor{blue!20}\textbf{0.669} & \cellcolor{blue!20}\textbf{13.99} \\
\bottomrule
\end{tabular}
}
\caption{Multi-benchmark evaluation of DeepSeek-MoE-16B under CIM hardware noise at \(T = 25\)°C (low noise) and \(T = 80\)°C (high noise). WT-2/LAM: perplexity (\(\downarrow\)); PIQA/ARC-C/ARC-E: accuracy (\(\uparrow\)). \textbf{Bold} indicates best performance under noise.}
\label{tab:deepseek_recovery}
\end{table}

\paragraph{Multi-Benchmark Evaluation on DeepSeek-MoE-16B}

Table~\ref{tab:deepseek_recovery} evaluates DeepSeek-MoE-16B across five benchmarks spanning language modeling (WikiText-2 and LAMBADA perplexity) and commonsense reasoning (PIQA, ARC-Easy, and ARC-Challenge accuracy) under two CIM noise regimes. ROMER achieves the best result on every benchmark at both temperatures.

Under vanilla deployment at \(T = 25\)°C, WikiText-2 perplexity rises from 18.94 to 23.83 (\(+4.89\)), LAMBADA from 10.64 to 14.88 (\(+4.24\)), and ARC-Challenge accuracy drops from 0.471 to 0.415 (\(-5.6\) points). At \(T = 80\)°C, degradations intensify dramatically: WikiText-2 perplexity surges to 28.36 (\(+9.42\)), LAMBADA to 19.53 (\(+8.89\)), and ARC-Challenge accuracy collapses to 0.368 (\(-10.3\) points)—a 21.9\% relative drop. This confirms that analog noise critically impairs MoE routing across both generative and reasoning tasks.

Bit-slicing and averaging offer mild improvements at \(T = 25\)°C but fail at high noise, with WikiText-2 perplexities of 25.61 and 26.92 at \(T = 80\)°C. K-B calibration performs comparably to or worse than vanilla, suggesting simple statistical recalibration is insufficient. Weight-remapping is the strongest baseline with WikiText-2 perplexity of 21.46 at \(T = 25\)°C and 23.86 at \(T = 80\)°C, but still leaves substantial gaps across all benchmarks.

At \(T = 25\)°C, ROMER attains WikiText-2 perplexity of 20.73, LAMBADA of 12.20, and ARC-Challenge accuracy of 0.450, reducing vanilla degradation by 3.10, 2.68, and 3.5 points respectively. At \(T = 80\)°C, ROMER achieves WikiText-2 perplexity of 22.48 and LAMBADA of 13.99, recovering approximately 62\% of vanilla degradation. Compared to weight-remapping, ROMER further reduces WikiText-2 perplexity by 1.38 and improves ARC-Challenge accuracy by \(+1.5\) points at high temperature, demonstrating that router-aware recalibration addresses root causes rather than compensating for weight perturbations. The larger relative degradation on reasoning tasks (21.9\% on ARC-Challenge and 49.7\% on WikiText-2) suggests noisy routing selectively suppresses expert specializations critical for multi-step reasoning.
\begin{figure}[!t]
\centering
\includegraphics[width=\columnwidth]{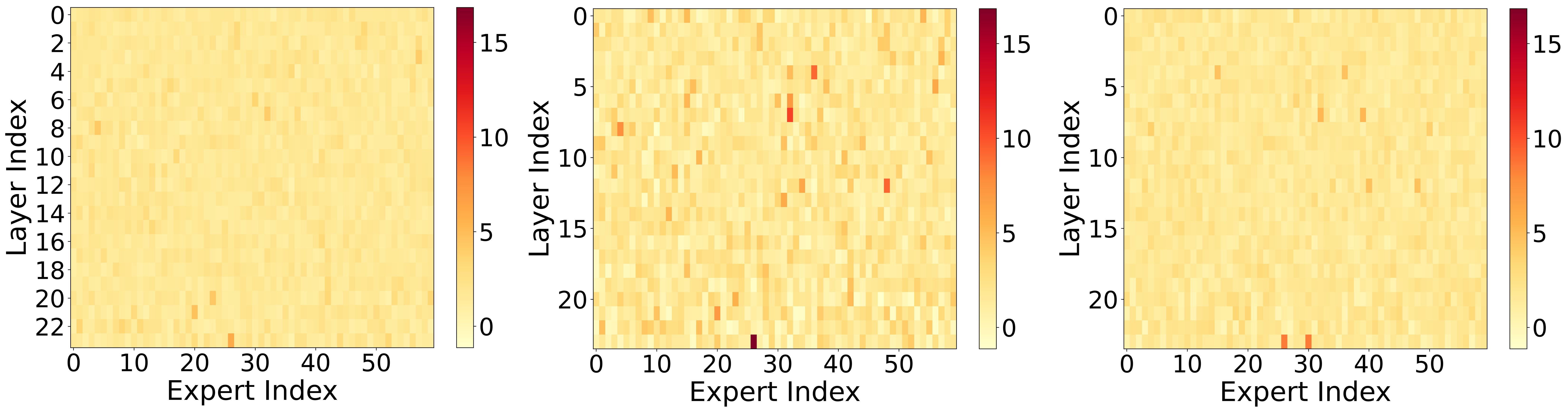}
\caption{Expert activation heatmaps of clean(left), vanilla (middle) and ROMER (right) under noise conditions for OLMoE-7B-A1B.}
\label{fig:recovery}
\end{figure}

\begin{table}[!t]
\renewcommand{\arraystretch}{1.3}
\centering
\resizebox{\linewidth}{!}{%
\begin{tabular}{l | r r c | r r c | r r c}
\toprule
\multirow{2}{*}{\textbf{Model}}
  & \multicolumn{3}{c|}{\textbf{Latency (s)} $\downarrow$}
  & \multicolumn{3}{c|}{\textbf{Power (W)} $\downarrow$}
  & \multicolumn{3}{c}{\textbf{Perplexity} $\downarrow$} \\
\cmidrule(lr){2-4}\cmidrule(lr){5-7}\cmidrule(lr){8-10}
  & Noisy & ROMER & $\Delta$
  & Noisy & ROMER & $\Delta$
  & Noisy & ROMER & $\Delta$ \\
\midrule
DeepSeek-MoE-16B
  & 245.3 & 261.0
  & \textcolor{blue}{$+$6.4\%}
  & 168.3 & 178.4
  & \textcolor{blue}{$+$6.0\%}
  & 35.0  & \textbf{14.48}
  & \textcolor{green!60!black}{$-$58.6\%} \\

Qwen1.5-MoE-A2.7B
  & 258.9 & 274.9
  & \textcolor{blue}{$+$6.2\%}
  & 179.1 & 190.2
  & \textcolor{blue}{$+$6.2\%}
  & 35.8  & \textbf{14.76}
  & \textcolor{green!60!black}{$-$58.8\%} \\

Qwen3-30B-A3B
  & 361.9 & 405.7
  & \textcolor{blue}{$+$12.1\%}
  & 245.8 & 273.6
  & \textcolor{blue}{$+$11.3\%}
  & 27.5  & \textbf{13.84}
  & \textcolor{green!60!black}{$-$49.7\%} \\

OLMoE-1B-7B
  & 234.6 & 249.8
  & \textcolor{blue}{$+$6.5\%}
  & 152.3 & 161.6
  & \textcolor{blue}{$+$6.1\%}
  & 35.3  & \textbf{14.20}
  & \textcolor{green!60!black}{$-$59.8\%} \\

\midrule
\textbf{Average} $\Delta$
  & \multicolumn{2}{r}{--}
  & \textbf{\textcolor{blue}{$+$7.8\%}}
  & \multicolumn{2}{r}{--}
  & \textbf{\textcolor{blue}{$+$7.4\%}}
  & \multicolumn{2}{r}{--}
  & \textbf{\textcolor{green!60!black}{$-$56.7\%}} \\
\bottomrule
\end{tabular}%
}
\caption{
    Hardware overhead and performance recovery of ROMER across four MoE models on a CIM hardware simulator at t = 85℃. \textit{Noisy}: vanilla deployment; \textit{ROMER}: calibrated deployment.
}
\label{tab:hardware_overhead}
\end{table}

\paragraph{Expert Activation Heatmap Analysis}

Figure~\ref{fig:recovery} visualizes expert activation frequency across all MoE layers under three conditions: clean inference, vanilla noisy deployment, and ROMER-calibrated inference. Each cell \((i, j)\) encodes the cumulative activation count of expert \(j\) in layer \(i\), with the colormap ranging from pale yellow (underactivated) to dark red (overactivated).

The clean model (left panel) exhibits broadly uniform activation across all layers and experts, reflecting intended MoE routing behavior where load is distributed evenly. This balanced pattern serves as the target for noise-robust methods.

Under CIM hardware noise (middle panel), the pattern becomes dramatically irregular with several experts showing disproportionately high activation counts (dark red cells) while large regions are rendered pale yellow, indicating severe expert underactivation. This routing collapse propagates across the full network depth, directly explaining the severe perplexity degradation in vanilla deployment as overloaded experts become bottlenecks while underactivated experts are wasted.

ROMER (right panel) substantially recovers the balanced activation distribution, eliminating overactivation hotspots and repopulating underactivation regions with moderate counts. This visual evidence corroborates quantitative results: by replacing underactivated experts and recalibrating router logits, ROMER targets the root causes of routing collapse and restores load-balanced behavior consistently across all layers.

\paragraph{Hardware Overhead Analysis}

Table~\ref{tab:hardware_overhead} reports the inference latency, power consumption, and perplexity of vanilla noisy deployment and ROMER-calibrated deployment across all four MoE models on CIM hardware. For three models (DeepSeek-MoE-16B, Qwen1.5-MoE-A2.7B, and OLMoE-1B-7B), ROMER introduces consistent overheads of approximately 6.2–6.5\% latency and 6.0–6.2\% power relative to vanilla noisy inference. This overhead stems from expert replacement, which redirects tokens to higher-frequency experts that may reside in different memory banks, incurring additional data movement. Since ROMER's calibration is performed entirely offline, the runtime overhead reflects only the modified routing pattern without additional computation.

Qwen3-30B-A3B exhibits higher overheads of +12.1\% latency and +11.3\% power due to its substantially larger active parameter count of 3B parameters per forward pass, compared to <1B for OLMoE and ~2.7B for Qwen1.5-MoE. A larger active footprint means remapping expert assignments incurs proportionally greater memory access overhead, amplifying the penalty. Despite this, ROMER achieves the lowest absolute perplexity (13.84) on Qwen3, confirming routing corrections remain highly effective at scale.

The key result demonstrates asymmetry between hardware cost and performance benefit. ROMER introduces average overheads of +7.8\% latency and +7.4\% power while achieving -56.7\% perplexity reduction relative to vanilla noisy deployment. Vanilla inference degrades perplexity to 27.5–35.8, rendering models essentially unusable, while ROMER recovers perplexity to 13.84–14.76, approaching clean baseline performance with single-digit percentage overhead for three of four models. This trade-off is practically acceptable in CIM deployments, where model quality degradation is the primary bottleneck rather than marginal increases in inference cost.
\begin{table}[!t]
\renewcommand{\arraystretch}{1.2}
\centering
\resizebox{\columnwidth}{!}{%
\begin{tabular}{lcccccc|cccccc}
\toprule
\multirow{2}{*}{\shortstack{Temperature\\℃}}
  & \multicolumn{6}{c|}{Expert Replacement}
  & \multicolumn{6}{c}{Router Calibration} \\
\cmidrule(lr){2-7} \cmidrule(lr){8-13}
 & \(n{=}0\) & \(n{=}2\) & \(n{=}4\) & \(n{=}6\) & \(n{=}8\) & \(n{=}16\)
 & \(\lambda{=}0\) & \(\lambda{=}0.2\) & \(\lambda{=}0.4\) & \(\lambda{=}0.6\) & \(\lambda{=}0.8\) & \(\lambda{=}1.0\) \\
\midrule
clean
  & \cellcolor{red!20}{10.83} & -- & -- & -- & -- & --
  & \cellcolor{red!20}{10.83} & -- & -- & -- & -- & -- \\
25
  & 11.61 & 11.05 & 11.01 & \cellcolor{blue!20}{\textbf{10.98}} & 11.41 & 25.44
  & 11.61 & 12.35 & \cellcolor{blue!20}{\textbf{10.40}} & 10.78 & 11.23 & 12.92 \\
45
  & 12.75 & 11.15 & \cellcolor{blue!20}{\textbf{10.89}} & 11.12 & 12.45 & 24.86
  & 12.75 & 13.24 & \cellcolor{blue!20}{\textbf{11.20}} & 11.63 & 12.11 & 13.87 \\
65
  & 15.57 & 13.49 & 12.12 & \cellcolor{blue!20}{\textbf{11.78}} & 13.15 & 24.12
  & 15.57 & 15.04 & \cellcolor{blue!20}{\textbf{13.20}} & 13.77 & 14.32 & 16.21 \\
85
  & 24.87 & 14.17 & 14.64 & \cellcolor{blue!20}{\textbf{14.05}} & 21.34 & 23.45
  & 24.87 & 22.64 & \cellcolor{blue!20}{\textbf{20.80}} & 21.72 & 22.45 & 24.10 \\
\bottomrule
\end{tabular}%
}
\caption{Ablation study of ROMER's two components on WikiText-2 perplexity (\(\downarrow\)) across five Temperature t, evaluated on Qwen3-30B-A3B.}
\label{tab:ablation_merged}
\end{table}

\paragraph{Ablation Study}

Table~\ref{tab:ablation_merged} ablates ROMER's expert replacement ($n$) and router calibration ($\lambda$) on WikiText-2 across Temperature $t \in \{25, 45, 65, 85\}$ using Qwen3-30B-A3B.
Expert replacement with $n{=}4$--6 substantially recovers performance. At $t{=}65$, $n{=}6$ reduces perplexity from 24.87 to 14.05 ($-10.82$). However, $n{\geq}8$ degrades performance, with $n{=}16$ worse than no replacement.
Router calibration at $\lambda{=}0.4$ consistently achieves lowest perplexity across all temperature. At $t{=}85$, it reduces perplexity to 20.80 ($-4.07$); at $t{=}25$, it achieves 10.40---below the clean baseline of 10.83. Both $\lambda{=}0.2$ and $\lambda{=}1.0$ degrade performance.
Both components address complementary failure modes and neither alone achieves ROMER's full recovery.

\section{Conclusion}

In this work, we propose ROMER, a novel training-free method to enhance the robustness of MoE-based LLMs on analog CIM hardware against inherent noise by replacing inactive experts with their active counterparts and adaptively adjusting router logits through percentile-based statistics. Extensive experiments across various datasets demonstrate that ROMER achieves up to 58.6\%, 58.8\%, and 59.8\% improvement under real-chip noise conditions for DeepSeek-MoE, Qwen-MoE, and OLMOE, respectively.


\clearpage

\section*{Ethics Statement}

We have adhered to the COLM Code of Ethics in this research. Our work is based entirely on publicly available models and benchmarks, and involves no human subjects.

\section*{Reproducibility Statement}

We are committed to the reproducibility of our work. All experiments were conducted on publicly available models and benchmarks.

\bibliography{colm2026_conference}
\bibliographystyle{colm2026_conference}

\clearpage

\end{document}